\pdfoutput=1
\documentclass[11pt]{article}
\usepackage[final]{acl}  

\usepackage{times}
\usepackage{latexsym}
\usepackage[T1]{fontenc}
\usepackage[utf8]{inputenc}
\usepackage{microtype}
\usepackage{inconsolata}
\usepackage{graphicx}

\usepackage{booktabs}
\usepackage{subcaption}
\usepackage{amssymb}
\usepackage{mathtools}
\usepackage{xspace}
\usepackage{bbm}
\usepackage{colortbl}
\usepackage{paralist}
\usepackage{xcolor}

\makeatletter
\DeclareRobustCommand\onedot{\futurelet\@let@token\@onedot}
\def\@onedot{\ifx\@let@token.\else.\null\fi\xspace}

\def\eg{\emph{e.g}\onedot}

\makeatother

\title{
Preserving Multi-Modal Capabilities of Pre-trained VLMs 

for Improving Vision-Linguistic Compositionality
}

\newcommand{\mailto}[2]{\href{mailto:#1}{\color{black}#2}}
\newcommand\coauthormark{\footnotemark[\arabic{footnote}]}
\author{%
    Youngtaek Oh\textsuperscript{1}\quad~ %
    Jae Won Cho\textsuperscript{2}\quad~ %
    Dong-Jin Kim\textsuperscript{3}\quad~ %
    In So Kweon\textsuperscript{1}\thanks{Corresponding authors}\quad~ %
    Junmo Kim\textsuperscript{1}\protect\coauthormark\vspace{2mm}\\
    \textsuperscript{1}KAIST \qquad\quad \textsuperscript{2}Sejong University \qquad\quad \textsuperscript{3}Hanyang University\vspace{1mm}\\%
        \textsuperscript{1}\texttt{%
            \{%
                \mailto{youngtaek.oh@kaist.ac.kr}{youngtaek.oh}, %
                \mailto{iskweon77@kaist.ac.kr}{iskweon77}, %
                \mailto{junmo.kim@kaist.ac.kr}{junmo.kim}%
            \}@kaist.ac.kr%
        }\\%
        \textsuperscript{2}\texttt{\mailto{chojw@sejong.ac.kr}{chojw@sejong.ac.kr}}\qquad%
        \textsuperscript{3}\texttt{\mailto{djdkim@hanyang.ac.kr}{djdkim@hanyang.ac.kr}}%
}%

\AtEndPreamble{
    \usepackage[capitalize]{cleveref}
    \crefname{section}{Sec.}{Secs.}
    \Crefname{section}{Section}{Sections}
    \Crefname{table}{Table}{Tables}
    \crefname{table}{Tab.}{Tabs.}
}

\begin{document}
\maketitle
\begin{abstract} 
In this paper, we propose a new method to enhance compositional understanding in pre-trained vision and language models (VLMs) without sacrificing performance in zero-shot multi-modal tasks. Traditional fine-tuning approaches often improve compositional reasoning at the cost of degrading multi-modal capabilities, primarily due to the use of global hard negative (HN) loss, which contrasts global representations of images and texts. This global HN loss pushes HN texts that are highly similar to the original ones, damaging the model's multi-modal representations. To overcome this limitation, we propose Fine-grained Selective Calibrated CLIP (\texttt{FSC-CLIP}), which integrates local hard negative loss and selective calibrated regularization. These innovations provide fine-grained negative supervision while preserving the model's representational integrity. Our extensive evaluations across diverse benchmarks for both compositionality and multi-modal tasks show that \texttt{FSC-CLIP} not only achieves compositionality on par with state-of-the-art models but also retains strong multi-modal capabilities. Code is available at: \url{https://github.com/ytaek-oh/fsc-clip}.
\end{abstract}

\section{Introduction}
\label{sec:1_intro}
\begin{figure}[t]
  \centering
   \includegraphics[width=0.99\columnwidth]{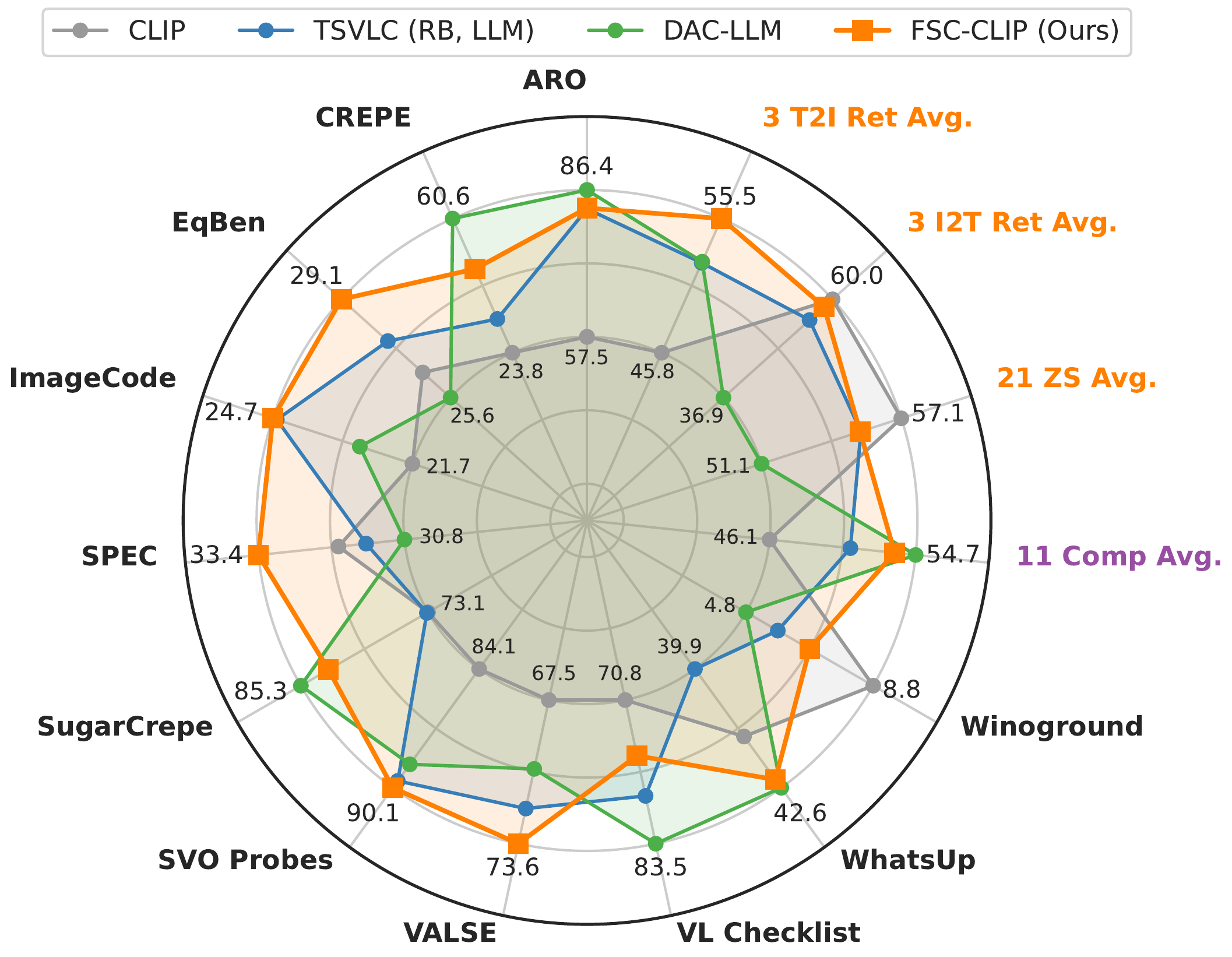}
   \vspace{-2mm}
   \caption{A holistic comparison of fine-tuning methods for vision-language compositionality. Enhancing compositionality often compromises multi-modal task performance in previous approaches. Our \texttt{FSC-CLIP} bridges this gap, minimizing these trade-offs. Full experimental results are provided in~\cref{tab:method_comparison}.
   }
   \label{fig:teaser}
   \vspace{-3mm}
\end{figure}
Humans naturally excel at multi-modal understanding, effortlessly perceiving and interpreting different modalities, such as images and text, and forming associations between them. This capability is evident in recognizing novel concepts~\cite{fu2018recent}, cross-modal retrieval~\cite{kaur2021comparative}, and compositional reasoning~\cite{levesque2012winograd}.
To achieve this ability in artificial intelligence, foundational vision and language models (VLMs) have been trained on large-scale image-text datasets~\cite{schuhmann2022laion}, significantly bridging the gap between human and machine capabilities in tasks like zero-shot recognition and image-text retrieval~\cite{radford2021learning}.

Despite these advances, VLMs still face challenges in compositional reasoning~\cite{yuksekgonul2023when}. 
Humans intuitively understand complex compositional language in combination with images, engaging in spatial reasoning~\cite{kamath-etal-2023-whats}, recognizing attributes and relationships in objects~\cite{hsieh2024sugarcrepe}, and perceiving equivariance between image and text~\cite{wang2023equivariant}.
In contrast, VLMs often fail to understand these nuanced relationships~\cite{liu2023visual,ray2024cola}.
This shortfall is attributed to their reliance on global, single vector representations~\cite{kamath-etal-2023-text} and limited ability to match compositional knowledge~\cite{wang2024diagnosing}.

To improve compositionality in VLMs, both pre-training~\cite{singh-etal-2023-coarse,zheng2024iterated} and fine-tuning~\cite{zhang2023contrasting,singh2024learn} methods have been proposed. In particular, fine-tuning, which leverages pre-trained knowledge and is cost-effective, is widely adopted in academia. Typically, this involves incorporating hard negative texts~\cite{doveh2022teaching,doveh2024dense,herzig-etal-2023-incorporating} into training. However, as shown in~\cref{fig:teaser}, this approach can result in a trade-off, where gains in compositionality come at the expense of performance in the multi-modal tasks: zero-shot classification (\texttt{ZS}) and image to text retrieval (\texttt{I2T Ret}). Previously, hard negative (HN) losses are applied to global image and text representations. Since HN texts are encoded too similarly to the original ones~\cite{kamath-etal-2023-text}, pushing them away with the HN loss can disrupt the multi-modal representations.

To address this, we propose a new fine-tuning framework for VLMs that enhances compositional reasoning while preserving performance in multi-modal tasks. Our approach mitigates the degradation caused by \underline{global hard negative loss} on \underline{single vector representations}, which struggles to capture subtle informational differences between hard negative texts and the original text. 

Our framework introduces two key innovations:
\noindent\textbf{(1) Local Hard Negative (LHN) Loss}. We utilize dense alignments between image patches and text tokens to calculate the hard negative loss. This approach, inspired by the dense alignment for vision-language representation~\cite{huang2021gloria,bica2024improving}, aggregates local similarity scores to enhance compositional understanding without undermining multi-modal representations.

\noindent\textbf{(2) Selective Calibrated Regularization (SCR)}.
To address the adverse effects of hard negative (HN) losses caused by similarly encoded HN and original texts, SCR is designed to better regulate HN supervision. It selectively focuses on challenging HN texts and applies a slight positive margin, reducing confusion and improving calibration.

The whole framework, dubbed \textbf{F}ine-grained \textbf{S}elective \textbf{C}alibrated CLIP, offers fine-grained supervision of hard negatives while preserving the integrity of multi-modal representations. As shown in~\cref{fig:teaser}, \texttt{FSC-CLIP} not only improves compositionality but also maintains high performance in multi-modal tasks. It outperforms DAC-LLM in \texttt{ZS} and \texttt{I2T Ret} scores, while achieving similar compositionality (\texttt{Comp}) across a wide range of tasks. We summarize our contributions as follows:
\vspace{-1mm}
{\setdefaultleftmargin{1mm}{}{}{}{}{}
\begin{itemize}
  \item We propose a novel fine-tuning methodology, \texttt{FSC-CLIP}, that aims to enhance vision-language compositionality in pre-trained VLMs while maintaining their multi-modal task capabilities.
  
  \item We design a local hard negative (LHN) loss and a selective calibrated regularization (SCR) mechanism, effectively capturing subtle differences in hard negative texts and preserving the integrity of multi-modal representations.
  
  \item We validate \texttt{FSC-CLIP} through an extensive range of experiments, covering 11 compositionality, 21 zero-shot recognition, and 3 image-text retrieval tasks, establishing a comprehensive evaluation of VLMs' multifaceted capabilities.
\end{itemize}
}

\section{Related Work}
\label{sec:4_related}
\textbf{Contrastive Vision-Language Models.}
CLIP \cite{radford2021learning} has revolutionized multi-modal domains through large-scale image-text pre-training, demonstrating remarkable zero-shot capabilities. Its dual encoder architecture has introduced versatility and driven advancements across a wide range of existing vision~\cite{zhou2022learning,oh2022daso,cho2022mcdal} and vision-language downstream tasks~\cite{jang2022sign,jang2023self,cho2023empirical,cho2023generative,cho2023counterfactual,kim-etal-2019-image,kim2021smem,kim2021dense}. CLIP also serves as the foundation for recognition~\cite{liang2023open},  image captioning~\cite{mokady2021clipcap,lee2024ifcap,kim2024semi,kim2024nice}, large multi-modal models~\cite{li2023blip_two,liu2024visual}, and generative models~\cite{podell2023sdxl}. In addition, CLIP extends its utility to connecting 3D~\cite{sun2023alpha} or audio~\cite{elizalde2023clap,senocak2023sound} to language, highlighting its essential role in multi-modal and compositional tasks in practical applications. We aim to enhance CLIP's compositional understanding while preserving its multi-modal capabilities.

\begin{figure*}[t]
  \centering
   \includegraphics[width=0.80\linewidth]{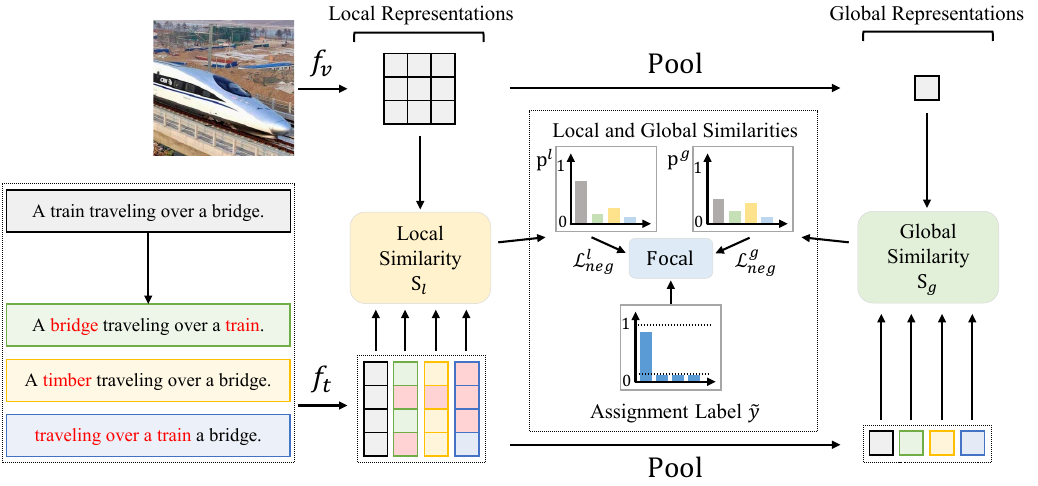}
   \vspace{-1mm}
   \caption{A complete \texttt{FSC-CLIP} framework that incorporates Local Hard Negative (LHN) Loss with Selective Calibrated Regularization (SCR), alongside a global HN loss. The LHN loss measures similarity between an image and a text at the patch and token levels to more accurately identify subtle differences between original and HN texts. SCR combines focal loss with label smoothing to mitigate the adverse effects of using hard negative losses.
   }
   \label{fig:overview}
   \vspace{-3mm}
\end{figure*}

\noindent \textbf{Vision-Language Compositionality.}
Although vision and language models exhibit promising capabilities such as zero-shot classification and retrieval~\cite{radford2021learning,pmlr-v162-zeng22c}, they still struggle with compositional reasoning, which requires fine-grained understanding between image and text~\cite{peng2023synthesize}. Numerous benchmarks have been proposed, testing various aspects like attributes, relationships and objects~\cite{zhao-etal-2022-explainable,yuksekgonul2023when}, spatial reasoning~\cite{kamath-etal-2023-whats,liu2023visual} and linguistic phenomena~\cite{parcalabescu-etal-2022-valse}. 
To enhance compositionality, incorporating hard negative captions during fine-tuning has become a common approach~\cite{zhang2023contrasting}, with these captions being generated through rule-based methods~\cite{doveh2022teaching,yuksekgonul2023when}, large language model prompting~\cite{doveh2024dense}, or scene graphs~\cite{singh-etal-2023-coarse,herzig-etal-2023-incorporating}.
We comprehensively evaluate the capabilities of VLMs across a broad range of compositionality and multi-modal tasks.

\section{Methodology}
\label{sec:2_method}
We first outline the fine-tuning setup for CLIP in~\cref{sec:2_1_clip}. Next, we introduce \texttt{FSC-CLIP}, which incorporates \textbf{Local Hard Negative (LHN) Loss} and \textbf{Selective Calibrated Regularization (SCR)} in \cref{sec:2_2_lhn,sec:2_3_scr}. The training objective for \texttt{FSC-CLIP} is described in \cref{sec:2_3_overall_obj}. 
The complete \texttt{FSC-CLIP} framework, integrating both global and local HN losses with SCR, is illustrated in~\cref{fig:overview}.

\subsection{CLIP with Global Contrastive Loss}
\label{sec:2_1_clip}
\noindent \textbf{CLIP objective.}
Consider a mini-batch $\mathcal{B}=\{ (I_{i},T_{i}) \}_{i=1}^{B}$ of size $B$, consisting of image and text pairs $(I_{i},T_{i})$. Using CLIP's visual and language encoders, $f_v(\cdot)$ (\eg, ViT~\cite{dosovitskiy2021an}) and $f_t(\cdot)$ (\eg, Transformers~\cite{vaswani2017attention}), each image $I_{i}$ is encoded into a sequence of visual tokens $\mathrm{\textbf{V}}_i = f_v(I_{i})$, and each text $T_{i}$ into a sequence of textual tokens $\mathrm{\textbf{T}}_i = f_t(T_{i})$. These sequences are represented in a shared multi-modal space, with $\mathrm{\textbf{V}}_i = \{ \mathrm{v}_{p,i} \}^{P}_{p=1}$ comprising $P$ patch embeddings and $\mathrm{\textbf{T}}_i = \{ \mathrm{t}_{w,i} \}_{w=1}^{W}$ consisting of $W$ token embeddings. The global representations of image and text $v_{i}$ and $t_{i} \in \mathbb{R}^d$ can be obtained by pooling the local representations: $v_{i}=\text{Pool}\left(\mathrm{\textbf{V}}_i \right)$  and $t_{i}=\text{Pool} \left(\mathrm{\textbf{T}}_i \right)$, respectively. For example, $\text{Pool}(\cdot)$ corresponds to \texttt{avgpool} and \texttt{argmax} for images and texts in~\citet{radford2021learning}.

CLIP aligns the corresponding images and texts by measuring the global-level similarity:
\begin{equation}
    \label{eqn:global_sim}
    \mathrm{S}_{g}\left(I_{i},T_{i} \right)= \exp \left(  
     \text{cos}\left(v_{i},t_{i} \right)/\tau \right),
\end{equation}
where $\text{cos}\left(v,t \right)= \frac{v^T t} {\lVert v \rVert \cdot \lVert t\rVert}$.
The image to text loss $\mathcal{L}_{i2t}$ of CLIP maximizes $\mathrm{S}_{g}\left(I_{i},T_{i} \right)$, while minimizing $\mathrm{S}_{g}\left(I_{i},T_{j} \right)$ for all non-matching texts $j \neq i$:
\begin{equation}
    \label{eqn:i2t_contra}
    \mathcal{L}_{i2t}= -\frac{1} {B} \sum_{i=1}^{B} \log \frac{\mathrm{S}_{g}\left(I_{i},T_{i} \right)} {\sum_{j=1}^{B} \mathrm{S}_{g}\left(I_{i},T_{j} \right) },
\end{equation}
and the text to image loss $\mathcal{L}_{t2i}$ is the reciprocal of $\mathcal{L}_{i2t}$ which aligns the matching image per text. The final CLIP loss is $\mathcal{L}_{\text{clip}}=\frac{1}{2} \left( \mathcal{L}_{i2t} + \mathcal{L}_{t2i}\right)$.

\noindent \textbf{Incorporating hard negative texts.} 
To enhance the compositional reasoning of CLIP, hard negative (HN) texts are commonly incorporated into training, whether they are rule-based~\cite{yuksekgonul2023when} or generated by language models~\cite{doveh2024dense}.
Consider a set of $K$ different HN texts  $\mathrm{\Tilde{T}}_{i}=\{\Tilde{T}_{i}^{k} \}^{K}_{k=1}$ originated from $T_{i}$. 
We introduce a separate hard negative loss added to $\mathcal{L}_{\text{clip}}$, similar to~\citet{doveh2022teaching}. First, we compute a similarity prediction probability $p^g_i$, assigned to the original caption $T_i$ as follows:
\begin{equation}
    \label{eqn:neg_contra}
    p^{g}_{i} = \frac{\mathrm{S}_{g}\left(I_{i},T_{i} \right)} {\mathrm{S}_{g}\left(I_{i},T_{i} \right) + \sum_{k=1}^{K} \mathrm{S}_{g}\left(I_{{i}},\Tilde{T}^{k}_{{i}} \right) }.
\end{equation}
Here, $g$ represents the global representation, and the hard negative (HN) loss applied to this similarity assignment is formulated as cross entropy:
\begin{equation}
    \label{eqn:neg_loss}
    \mathcal{L}^{{g}}_{neg}=-\frac{1}{B}\sum^{B}_{i=1}\log p^{g}_{i}.  
\end{equation}

However, incorporating such global HN loss can inadvertently harm the multi-modal representations due to the similarly encoded global text representations between original and HN texts.

\subsection{Local Hard Negative (LHN) Loss}
\label{sec:2_2_lhn}
To address this, we propose a novel Local Hard Negative (LHN) loss that utilizes a local similarity score $\mathrm{S}_l(I, T)$. Replacing the global similarity $\mathrm{S}_g$ with $\mathrm{S}_l$, the LHN loss is formulated as follows:
\begin{equation}
    \label{eqn:local_neg}
    \mathcal{L}^{{l}}_{neg} =  \frac{-1}{B} \sum^B_{i=1} \log \frac{\mathrm{S}_l \left(I_i, T_i \right)} {\underbrace{ \mathrm{S}_l \left(I_i, T_i \right) + \displaystyle\sum^K_{k=1} \mathrm{S}_l \left(I_i, \Tilde{T}^k_i \right)}_{ p^{l}_{i}} },
\end{equation}
where $p^l_i$ represents the local similarity prediction.

Unlike \citet{bica2024improving}, which uses token-level contrast for image-text pairs, we introduce a new HN loss based on local similarity $S_l$ from token-patch representations, enabling the capture of subtle differences between the original and HN texts.

\noindent \textbf{Textual-aligned Visual Patches.}
$\mathrm{S}_l(I, T)$ is designed to measure the similarity between token and patch embeddings for each token in the given text $T$. From the patch representations $\mathrm{\textbf{V}}=\{\mathrm{v}_p \}^P_{p=1}$, we first derive the textual-aligned patch embeddings $\mathrm{\hat{\textbf{V}}} = \{ \hat{\mathrm{v}}_w \}_{w=1}^W$, corresponding to each textual token feature $\mathrm{t}_w$ in $\mathrm{\textbf{T}}\in \mathbb{R}^{W\times d}$. 
This is achieved by performing a weighted average of patches $\mathrm{\textbf{V}}$ using attention weights $\mathrm{a} \in \mathbb{R}^{W\times P}$ derived from normalizing the similarity map $\mathrm{s}$ between token and patch embeddings. 
We denote the similarity map as $\mathrm{s}=\mathrm{\textbf{T}}^T \mathrm{\textbf{V}} \in \mathbb{R}^{W\times P}$, where $\mathrm{s}_{w, p}=\mathrm{t}^T_w \mathrm{v}_p$.  

To relate multiple similar patches for each token, we min-max normalize $\mathrm{s}$ to obtain $\mathrm{a}$:
\begin{equation}
    \label{eqn:attn_norm}
    \mathrm{a}_{w,p}=\frac{\mathrm{s}_{w, p} - \min_k {\mathrm{s}_{w, k}}} {\max_k {\mathrm{s}_{w, k}} - \min_k {\mathrm{s}_{w, k}}},
\end{equation}
and use the attention weights $\mathrm{a}$ to aggregate $\mathrm{\textbf{V}}$, obtaining the textual-aligned patches $\mathrm{\hat{\textbf{V}}}=\{\hat{\mathrm{v}}_w \}^W_{w=1}$:
\begin{equation}
    \hat{\mathrm{v}}_w = \frac{1}{\sum^P_{p=1} \mathrm{a}_{w, p}} \cdot \sum^{P}_{p=1} \mathrm{a}_{w,p} \cdot \mathrm{v}_p. 
\end{equation}
In \cref{sec:sup_additional_analysis}, we explore different normalization choices for the attention weights in~\cref{eqn:attn_norm}.

\noindent \textbf{Token-level Similarity.} 
After obtaining the textual-aligned visual tokens $\hat{\mathrm{\textbf{V}}}$, we aggregate the per-token similarities between $\hat{\mathrm{\textbf{V}}}$ and $\mathrm{\textbf{T}}$ as follows:
\begin{equation}
    \mathrm{S}_l \left(I, T \right) = \sum^W_{w=1} \exp \left(\cos \left(
    \mathrm{\hat{v}}_w, \mathrm{t}_w \right)/\tau \right),
\end{equation}
where $\hat{\mathrm{v}}_w \in \mathbb{\hat{\mathrm{\textbf{V}}}}$ and $ \mathrm{t}_w \in \mathrm{\textbf{T}}$. 
Unlike $\mathrm{S}_g(I, T)$ which is based on global representations, $\mathrm{S}_l (I, T)$ focuses on the local alignment between image and text, better distinguishing features between correct and HN texts, thereby reducing the negative impact by the hard negative loss, as illustrated in~\cref{fig:overview}.

We observe that $\mathcal{L}^{{l}}_{neg}$ maintains multi-modal task performance close to the pre-trained representations while significantly enhancing compositionality.  
Notably, the order of aggregation, whether pooling first and then computing similarity (\eg, $S_g$), or computing token-level similarity before aggregation (\eg, $S_l$), proves to be important.

\begin{figure}[t]
  \centering
   \includegraphics[width=0.85\columnwidth]{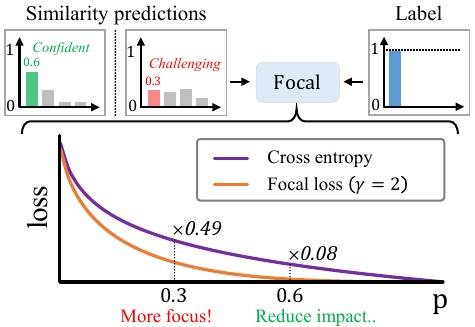}
   \vspace{-1mm}
   \caption{A conceptual illustration of the confidence-based weighting mechanism in HN loss. It reduces the adverse impact of HN supervision by lowering the signal from confident predictions while selectively focusing on challenging ones, crucial for learning compositionality.
   }
   \label{fig:focal_mechanism}
   \vspace{-3mm}
\end{figure}
\subsection{Selective Calibrated Regularization (SCR)}
\label{sec:2_3_scr}
Since hard negative (HN) texts are often encoded similarly to the original texts, HN losses can disrupt multi-modal representations. To counter this, we propose Selective Calibrated Regularization (SCR) to better regulate HN supervision, seamlessly applicable to both global and local HN losses. 

SCR has two components: one modulates the supervision signal based on image-text similarity, while the other adjusts label assignments to calibrate the positiveness of HN texts. As shown in~\cref{tab:compo_analysis}, we confirm that both components are crucial for preserving the representation integrity.

\noindent \textbf{Focal Loss to Target Challenging HN Texts.}
To mitigate the negative impact of supervising HN texts, we reduce the supervision signal for confident similarity predictions to the original text. Instead, we focus more on challenging HN texts that exhibit higher similarity to the image and may be confused with the original texts. This confidence-based weighting aligns with the concept of focal loss~\cite{lin2017focal}, as shown in~\cref{fig:focal_mechanism}.

Formally, let the similarity prediction for the $i$-th batch item, including $K$ generated HN texts, be represented as a vector $\mathrm{p}_i \in \mathbb{R}^{1+K}$, where the first element corresponds to the original text. 
The HN loss can be re-formulated in a vector representation with $\mathrm{p}_i$ as 
$\text{CE}(\mathrm{p}_i, \mathrm{y}_i)= \sum^{K}_{k=0} l_{i,k}$, where $l_{i,k}= -\mathrm{y}_{i, k}\log \mathrm{p}_{i,k}$ and $\mathrm{y}_i= \mathbbm{1}_{\left[k=0\right]} \in \mathbb{R}^{1+K}$ indicates the assignment label between an image and all texts. To reduce the negative impact of the confident image-text similarity predictions, we apply confidence-based weighting to $\text{CE}$ loss as follows:
\begin{equation}
    \label{eqn:focal}
    \text{Focal}\left(\textrm{p}_{i}, \textrm{y}_{i} \right) = \sum^{K}_{k=0} \left(1-\mathrm{p}_{i,k} \right)^{\gamma} l_{i,k}, 
\end{equation}
where $\gamma$ is the modulation parameter. 
This strategy prioritizes challenging image-text associations, essential for learning compositionality, while effectively preventing degradation from the HN loss.

\noindent \textbf{Label Smoothing to Calibrate the Positiveness of HN Texts.}
Although hard negative (HN) texts share similar representations with the original text, previous methods have overlooked their potential positiveness in the HN loss design, assigning a strict value of 0 to all HN texts in the label vector $\mathrm{y}_i$. Similar to the motivation in~\citet{zhang2023contrasting}, but differing from their ranking loss approach, we acknowledge the potential correctness of HN texts by assigning a slight positive margin rather than categorizing them as entirely negative.

To this end, we apply label smoothing~\cite{guo2017calibration} to the label vector $\mathrm{y}_i$ using a smoothing parameter $\beta$ to ensure a positive margin for HN texts:
\begin{equation}
    \Tilde{\mathrm{y}}_{i,k}= (1-\beta)\cdot \mathrm{y}_{i,k} + \frac{\beta}{1+K}, 
\end{equation}
where $\Tilde{\mathrm{y}}_i$ provides a non-binary label for the HN losses. 
It helps preserve the model's representations during training with HN losses.

\subsection{Overall Training Objective}
\label{sec:2_3_overall_obj}
Our \texttt{FSC-CLIP} incorporates two hard negative (HN) losses, $\mathcal{L}^{{g}}_{neg}$ and $\mathcal{L}^{{l}}_{neg}$, representing global and local HN losses respectively, into CLIP  loss~$\mathcal{L}_{\text{clip}}$:
\begin{equation}
    \mathcal{L}_{\text{total}}=\mathcal{L}_{\text{clip}}+\lambda_g\mathcal{L}^{{g}}_{neg} + \lambda_l \mathcal{L}^{{l}}_{neg},
\end{equation}
where $\lambda_g$ and $\lambda_l$ are the weighting factors for the respective losses. Selective Calibrated Regularization (SCR) is applied to both losses, incorporating label smoothing and focal loss. The global HN loss, $\mathcal{L}^{{g}}_{neg}$ is computed as $\text{Focal}\left(\mathrm{p}^g, \Tilde{\mathrm{y}} \right)$, while the LHN loss, $\mathcal{L}^{{l}}_{neg}$ is derived similarly, by replacing $\mathrm{p}^g$ with $\mathrm{p}^l$ for the local representations.

\definecolor{mygray}{gray}{0.6}
\newcommand*\rot{\rotatebox{90}}
\begin{table*}[t]
\centering
\resizebox{.99\linewidth}{!}{%
    \begin{tabular}{lcccccccccccccccc}
    \toprule
     Method & LoRA                                        &  \rot{ARO}  &  \rot{CREPE}  &  \rot{EqBen}  &  \rot{ImageCoDe}  &  \rot{SugarCrepe}  &  \rot{SVO Probes}  &  \rot{VALSE}  &  \rot{VL-Checklist~~}  &  \rot{WhatsUp}  &  \rot{Winoground}  &  \rot{SPEC}  &  \rot{\texttt{Comp}}  &  \rot{\texttt{ZS}}  &  \rot{\texttt{I2T Ret}}  &  \rot{\texttt{T2I Ret}}  \\
    \midrule
CLIP (ViT-B/32)  &  & 57.5  &  23.8   &  26.5   &    21.7     &     73.1     &     84.1     &  67.5   &      70.8      &   41.5    &     8.8      &  31.9  &    46.1    &      57.1      &   60.0    &   45.8    \\
\cmidrule(lr){1-2} \cmidrule(lr){3-13} \cmidrule(lr){14-17}
\multicolumn{17}{c}{\textit{Fine-tuned: MS-COCO, 100K Samples}} \\
NegCLIP$^1$    &  & 80.9  &  30.3   &  \textbf{30.3}   &    \underline{26.4}     &     83.7     &     \underline{90.8}     &  73.7   &      74.9      &   \textbf{42.9}    &     8.0      &  \textbf{34.6}  &    52.4    &      \underline{55.9}      &   \textcolor{mygray}{66.8}    &   \textcolor{mygray}{58.4}    \\
CE-CLIP$^2$  &     & 76.3  &  34.7   &  26.8   &    24.5     &     \textbf{85.7}     &     90.1     &  \textbf{76.7}   &      \underline{76.9}      &   41.7    &     5.2      &  33.0  &    52.0    &      49.9      &   \textcolor{mygray}{59.2}    &   \textcolor{mygray}{57.4}    \\
GNM-CLIP$^3$   &  & 57.1  &  17.4   &  28.3   &    25.0     &     78.7     &     89.2     &  71.1   &      70.6      &   \underline{42.1}    &     \textbf{10.2}     &  33.1  &    47.5    &      \textbf{56.3}      &   \textcolor{mygray}{66.1}    &   \textcolor{mygray}{55.5}    \\
MosaiCLIP$^{\dagger,4}$ & & 82.6 & - & - & - & - &  90.7 & - & {76.8} & - & - & - & - & - & - & - \\
\cmidrule(lr){1-2} \cmidrule(lr){3-13} \cmidrule(lr){14-17}
NegCLIP$^\ddagger$ & & 85.0  &  34.7   &  \underline{29.8}   &    26.2     &     84.5     &     90.6     &  74.7   &      75.4      &   41.2    &     8.2      &  \underline{34.2}  &    53.1    &      {55.1}      &   \textcolor{mygray}{66.1}    &   \textcolor{mygray}{57.9}    \\
\textbf{\texttt{FSC-CLIP} (Ours)} &      & \underline{85.1}  &  \underline{42.2}   &  \underline{29.8}   &    26.3     &     \underline{85.1}     &     \textbf{90.9}     &  \underline{75.3}   &      76.7      &   40.6    &     \underline{9.5}      &  \underline{34.2}  &    \textbf{54.2}    &      {55.7}      &   \textcolor{mygray}{66.3}    &   \textcolor{mygray}{58.3}    \\
\textbf{\texttt{FSC-CLIP} (Ours)} & \checkmark & \textbf{85.2}  &  \textbf{42.9}   &  29.7   &    \textbf{26.5}     &     82.1     &     90.4     &  75.0   &      \textbf{77.2}      &   41.7    &     6.0      &  33.2  &    \underline{53.6}    &      {55.6}      &   \textcolor{mygray}{65.3}    &   \textcolor{mygray}{57.2}    \\

\cmidrule(lr){1-17}

\multicolumn{17}{c}{\textit{Fine-tuned: Conceptual Captions - 3M (CC-3M), \textbf{3M} Samples}} \\
TSVLC$^5$ (RB)  & \checkmark & 83.5  &  36.1   &  27.4   &    24.0     &     76.9     &     \textbf{89.8}     &  69.3   &      77.5      &   40.9    &     6.8      &  31.6  &    51.2    &      \underline{54.9}      &   54.9    &   52.1    \\
TSVLC$^5$ (RB+LLM) & \checkmark & 82.7  &  33.1   &  \underline{27.6}   &    \underline{24.6}     &     73.2     &     \underline{89.7}     &  72.2   &      79.2      &   39.9    &     5.8      &  31.4  &    50.9    &      \textbf{55.4}      &   55.1    &   52.3    \\
DAC-LLM$^6$  & \checkmark & \underline{86.4}  &  \underline{60.6}   &  25.6   &    22.8     &     \textbf{85.3}     &     88.9     &  70.5   &      \underline{83.5}      &   \underline{42.6}    &     4.8      &  30.8  &    \underline{54.7}    &      51.1      &   36.9    &   52.4    \\
DAC-SAM$^6$  & \checkmark & 83.3  &  \textbf{63.7}   &  25.3   &    24.3     &     83.8     &     88.5     &  70.2   &      \textbf{84.7}      &   42.4    &     \textbf{8.5}      &  29.9  &    \textbf{55.0}    &      51.9      &   41.1    &   49.0    \\
MosaiCLIP$^{\dagger,4}$ & & 80.4 &  - & - & - & - & - & - & 77.3 & - & - & - & - & 53.5 & - & - \\

\cmidrule(lr){1-2} \cmidrule(lr){3-13} \cmidrule(lr){14-17}

\multicolumn{17}{c}{\textit{Fine-tuned: Conceptual Captions -- 3M (CC-3M), \textbf{100K} Samples}} \\
NegCLIP$^\ddagger$ &  & \textbf{86.5}  &  50.5   &  25.8   &    \underline{24.6}     &     83.4     &     88.6     &  72.4   &      79.0      &   \textbf{43.2}    &     \underline{7.0}      &  \underline{32.8}  &    54.0    &      52.6      &   51.8    &   \underline{54.1} \\
\textbf{\texttt{FSC-CLIP} (Ours)} &  & 78.8 & 44.0 & \textbf{28.5} & \textbf{25.2} & \underline{84.3} & 88.2 & \textbf{74.9} & 77.4 & \underline{42.6} & 6.8 & \textbf{34.2} & 53.2 & 53.5 & \underline{55.8} & \textbf{54.6} \\
\textbf{\texttt{FSC-CLIP} (Ours)} & \checkmark & 84.4 & 50.6 & 27.7 & 24.5 & 82.3 & 88.8 & \underline{74.5} & 80.3 & 42.1 & 5.0 & 32.2 & 53.9 & 53.6 & \textbf{56.1} & 54.0 \\

\cmidrule(lr){1-17}

\multicolumn{17}{c}{\textit{Fine-tuned: LAION-COCO, \textbf{600M} Samples}} \\
CLoVe$^7$  &  & 83.0  &  41.7   &  26.9   &    \textbf{25.3}     &     \textbf{84.6}     &     87.9     &  71.8   &      66.6      &   41.8    &     6.5      &  31.7  &    51.6    &      51.0      &   53.1    &   \textbf{56.0}    \\
\cmidrule(lr){1-2} \cmidrule(lr){3-13} \cmidrule(lr){14-17}
\multicolumn{17}{c}{\textit{Fine-tuned: LAION-COCO, \textbf{100K} Samples}} \\
 NegCLIP$^\ddagger$ & & \textbf{86.4}  &  \underline{48.7}   &  \underline{27.2}   &    \textbf{25.3}     &     80.9     &     89.6     &  70.9   &      \underline{76.0}      &   \textbf{43.0}    &     \textbf{7.8}      &  32.3  &    \underline{53.5}    &      54.1      &   52.3    &   54.1    \\
 
\textbf{\texttt{FSC-CLIP} (Ours)} &      & 82.8  &  46.8   &  \textbf{29.1}   &    24.7     &     \underline{82.6}     &     \textbf{90.1}     &  \textbf{73.6}   &      75.7      &   42.4    &     \underline{6.8}      &  \textbf{33.4}  &    \underline{53.5}    &      \underline{55.3}      &   \textbf{58.2}    &   \underline{55.5}    \\
\textbf{\texttt{FSC-CLIP} (Ours)} & \checkmark & \underline{85.5}  &  \textbf{54.4}   &  \textbf{29.1}   &    \underline{24.9}     &     80.6     &     \underline{89.7}     &  \underline{72.6}   &      \textbf{78.4}      &   \underline{42.8}    &     5.8      &  \underline{32.5}  &    \textbf{54.2}    & \textbf{55.9} &   \underline{57.3}    &   {54.3}    \\
    \bottomrule
    \multicolumn{17}{l}{\normalsize $^\dagger$Numbers taken from the original paper.  ~$^\ddagger$Our implementation, without additional image batch.} \\
    \multicolumn{17}{l}{\small 
        References: $^1$\cite{yuksekgonul2023when} $^2$\cite{zhang2023contrasting} $^3$\cite{sahin2024enhancing} $^4$\cite{singh-etal-2023-coarse} $^{5,6}$\cite{doveh2022teaching,doveh2024dense} $^7$\cite{castro2024clove}
    } \\
\end{tabular}%
}
\vspace{-1mm}
\caption{A holistic comparison of fine-tuning methods applied to the pre-trained CLIP ViT-B/32 model across 11 compositionality, 21 zero-shot classification, and 3 retrieval tasks, including their meta averages: \texttt{Comp}, \texttt{ZS}, and \texttt{I2T/T2I Ret}. \texttt{FSC-CLIP} achieves superior compositionality scores while maintaining strong multi-modal task performances. For each fine-tuning dataset, the best numbers are \textbf{bold}, and the second-best numbers are \underline{underlined}.}
\label{tab:method_comparison}
\vspace{-3mm}
\end{table*}
\section{Experiments}
\label{sec:3_exp}
\noindent \textbf{Training Datasets.}
We consider three image-text datasets for fine-tuning: COCO captions~\cite{chen2015microsoft}, CC-3M~\cite{sharma-etal-2018-conceptual}, and LAION-COCO~\cite{Schuhmann2022laioncoco}. For COCO captions, we utilize 100K examples pre-processed by~\citet{yuksekgonul2023when}. As pointed out by~\citet{singh-etal-2023-coarse}, COCO shares data with several evaluation benchmarks (\eg, SugarCrepe and retrieval tasks), which may inadvertently affect the results. To ensure a broader evaluation and avoid such overlap, we additionally consider CC-3M and LAION-COCO for fine-tuning. For each dataset, we randomly sample 100K examples and, instead of using raw captions, we utilize synthetic captions paired with images. Specifically, for CC-3M, we generate captions using CoCa~\cite{yu2022coca} with ViT-L/14, while for LAION-COCO, we use captions generated by BLIP~\cite{li2022blip} with ViT-L/14, applied to the LAION-2B dataset~\cite{schuhmann2022laion}. 

\noindent \textbf{Hard Negative (HN) Texts.}
We employ simple rule-based methods for generating hard negative (HN) texts, avoiding the need for external language models like~\citet{le2023bloom} used in~\citet{doveh2024dense}. For each original caption, we apply three distinct operations: \texttt{negclip}, \texttt{replace}, and \texttt{bi-gram shuffle}. These operations are applied at every training step, ensuring variation in HN texts across iterations. As a result, each batch item is paired with an image and four captions, as illustrated in~\cref{fig:overview}. Further details and examples on these operations are provided in~\cref{sec:sup_neg_texts}.

\noindent \textbf{Training Setup.}
Consistent with previous methods~\cite{yuksekgonul2023when,singh-etal-2023-coarse,zhang2023contrasting}, we trained our models during 5 epochs with batch size 256, using OpenCLIP repository~\cite{ilharco_gabriel_2021_5143773}. 
The learning rate is set to 5e-6 and decayed by a cosine schedule, with a warmup of 50 steps. Models are optimized using AdamW with a weight decay of 0.1. We use a single Quadro RTX 8000 GPU with 48GB memory for training. Images are re-scaled to 224, and the context length is 77 for texts. We set the weighting factors $\lambda_g=0.5$ and $\lambda_l=0.2$. For SCR, we set $\gamma=2.0$ and $\beta=0.02$ for focal loss and label smoothing, respectively. We also experiment with LoRA~\cite{hu2022lora}, which preserves the original model parameters. Consistent with~\citet{doveh2022teaching,doveh2024dense}, we set the rank to 4.
Training our model takes less than one hour for 100K samples.

\noindent \textbf{Evaluation Setup.}
We utilize an \textit{extensive} range of benchmarks for a comprehensive evaluation, exceeding the scope of previous works. Full details including references are provided in~\cref{sec:sup_details_eval}. 

For compositionality, we employ 11 benchmarks in total: ARO, CREPE-Productivity, EqBen, ImageCoDe, SPEC, SugarCrepe, SVO Probes, VALSE, VL-Checklist, WhatsUp, and Winoground, testing different facets of compositional reasoning.
For multi-modal tasks, we evaluate 21 zero-shot classification tasks using ELEVATER toolkit. In addition, we conduct image-text retrieval evaluations on COCO, Flickr30k, and COCO-Counterfactuals. All those evaluations are performed using the \texttt{vl-compo} package~\cite{oh2024exploring}.

We report a single aggregated number, which is the average of sub-tasks for each compositionality benchmark. We also provide the meta-average across all compositionality benchmarks (\texttt{Comp}), the average performance over 21 zero-shot classification tasks (\texttt{ZS}), and the average Recall@1 for three image to text (\texttt{I2T Ret}) and text to image (\texttt{T2I Ret}) retrieval tasks, as shown in~\cref{tab:method_comparison}. 
For a fair comparison, we consistently run evaluations for all the previous models across all the benchmarks.

\subsection{Main Results}
\begin{figure}[t]
  \centering
   \includegraphics[width=0.99\columnwidth]{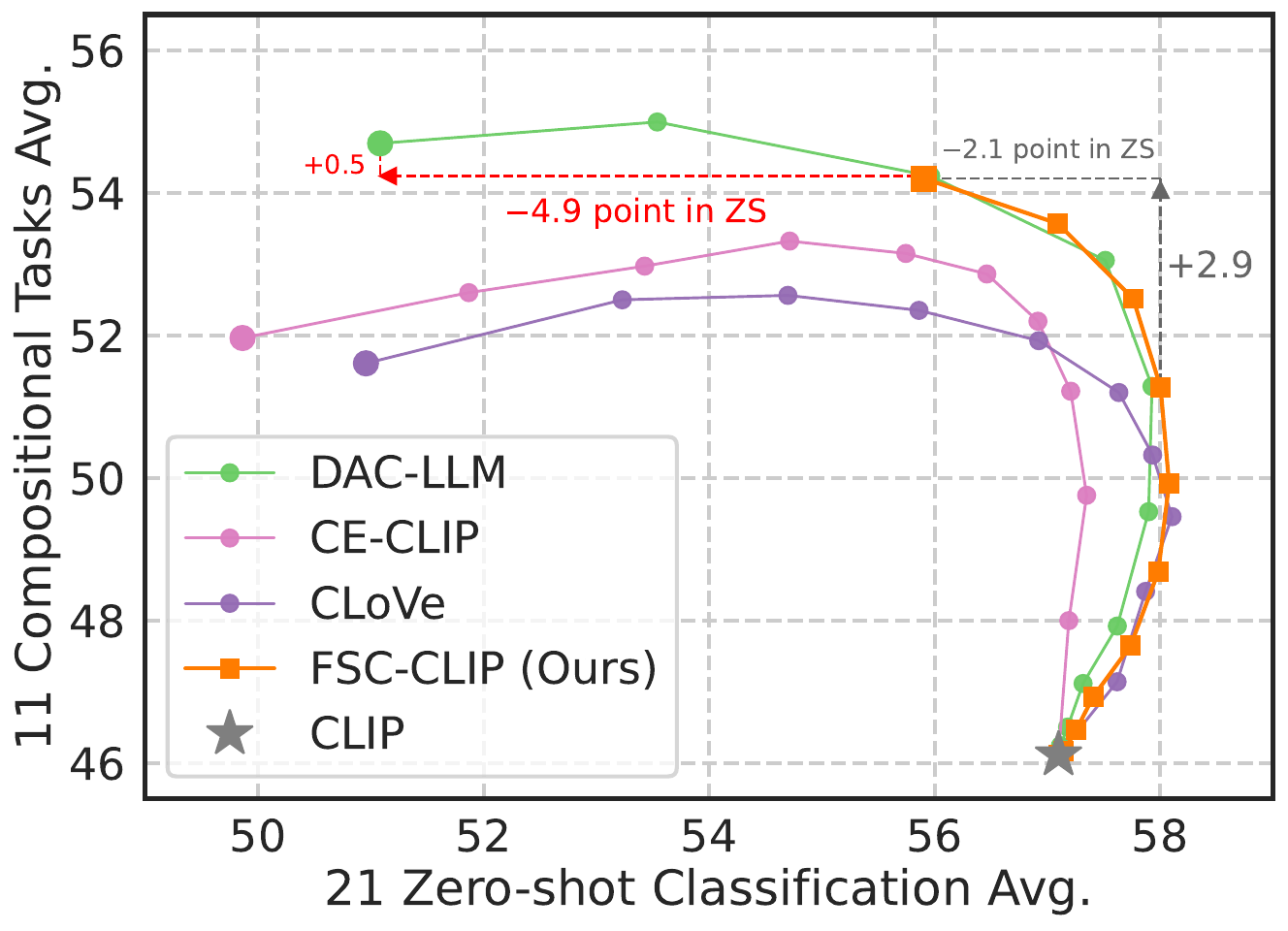}
   \caption{Fine-tuning trajectories between compositionality (\texttt{Comp}) and zero-shot classification (\texttt{ZS}) via robust fine-tuning method~\cite{wortsman2022robust}. 
   Each point represents the interpolated model between the pre-trained and each fine-tuned version, at varying ratios.
   \texttt{FSC-CLIP} offers better trade-offs between \texttt{Comp} and \texttt{ZS}, maintaining \texttt{ZS} scores in the fully fine-tuned model.
   } %
   \label{fig:wiseft}
   \vspace{-4mm}
\end{figure}

We compare our \texttt{FSC-CLIP} to previous fine-tuning methods for compositionality. We report both compositionality and multi-modal task performance as shown in~\cref{tab:method_comparison}.
In~\cref{fig:wiseft}, we visualize the trade-off trajectory between \texttt{Comp} and \texttt{ZS} through the robust fine-tuning method~\cite{wortsman2022robust}.

\noindent \textbf{Compositionality while Sacrificing Multi-Modal Tasks.}
We introduce our baseline, NegCLIP$^\ddagger$, serving as a direct comparison to our \texttt{FSC-CLIP}. Unlike the original implementation of NegCLIP~\cite{yuksekgonul2023when}, we utilize an online version of hard negatives generation (\eg, \texttt{negclip}) and omit the use of additional similar image batches. This baseline will be further used in our ablation study, with the symbol $^\ddagger$ omitted for convenience.

As shown in~\cref{tab:method_comparison}, we first compare our \texttt{FSC-CLIP} with previous models fine-tuned on COCO, aligning our results with those in the literature. 
CE-CLIP$^2$ shows a significant drop in \texttt{ZS} score to 49.9. Meanwhile, GNM-CLIP$^3$ maintains a \texttt{ZS} score close to that of the pre-trained model, but shows only a modest increase in \texttt{Comp}. In contrast, our model achieves superior \texttt{Comp} scores while maintaining competitive \texttt{ZS} and retrieval performance. As note, we have grayed out the retrieval scores of models fine-tuned on COCO to account for the influence of overlapping data.

When fine-tuned on datasets other than COCO, such as CC-3M and LAION-COCO, all baseline models show improvements in the \texttt{Comp} score, but this comes at the expense of their \texttt{ZS} and \texttt{I2T Ret} scores compared to the pre-trained CLIP. For example, NegCLIP$^\ddagger$ demonstrates promising \texttt{Comp} scores compared to methods like TSVLC$^5$ and CLoVe$^7$, but still shows weaker \texttt{ZS} and \texttt{I2T Ret} scores relative to the pre-trained model. Similarly, DAC-LLM$^6$, despite having the strongest \texttt{Comp} score supported by LLM-augmented captions, suffers notable declines in both \texttt{ZS} and \texttt{I2T Ret}, decreasing by 6.0 and 23.1 points, respectively. Although TSVLC$^5$ preserves these scores better than other models, its \texttt{Comp} score improvements are relatively smaller.
These methods apply hard negative (HN) loss to global-level representations, potentially causing the observed performance drops.

\begin{table}[t]
\centering
\resizebox{.99\linewidth}{!}{%
\begin{tabular}{ccccccccc}
\toprule
id & $\mathcal{L}^{{g}}_{neg}$ & $\mathcal{L}^{{l}}_{neg}$ & Focal & LS &  \texttt{Comp}  &  \texttt{ZS}  &  \texttt{I2T Ret}  &  \texttt{T2I Ret}  \\
\midrule
1 & \checkmark & - & - & - &    54.0    &      53.6      &   47.4    &   53.7    \\
2 & - & \checkmark & - & - & 51.7    &      55.7      &   61.6    &   54.5    \\
3 & \checkmark & \checkmark & - & - & 54.4    &      52.6      &   46.9    &   53.8    \\
\cmidrule(lr){1-5} \cmidrule(lr){6-9}  
4 & \checkmark & \checkmark & \checkmark & - & 54.2    &      54.2      &   53.1    &   54.8    \\
5 & \checkmark & \checkmark & - & \checkmark & 53.9    &      53.8      &   51.7    &   54.9    \\
6 & \checkmark & \checkmark & \checkmark & \checkmark & 53.5    &      55.3      &   58.2    &   55.5    \\
\cmidrule(lr){1-5} \cmidrule(lr){6-9}  
7 & \checkmark & - & \checkmark & \checkmark &  52.8    &      55.3      &   57.1    &   55.6    \\
8 & - & \checkmark & \checkmark & \checkmark & 50.2    &      55.9      &   63.2    &   55.1    \\
\bottomrule
\end{tabular}%
}
\caption{Impact by individual component. The local HN loss preserves multi-modal task performance. In addition, focal loss and label smoothing (LS) in SCR complement each other, improving the decreased multi-modal task performance caused by the HN losses.}
\vspace{-4mm}
\label{tab:compo_analysis}
\end{table}
\noindent \textbf{Preserving Multi-Modal Tasks.}
\texttt{FSC-CLIP} stands out by achieving higher \texttt{Comp} scores than previous models, comparable to DAC-LLM, while maintaining strong performance in multi-modal tasks. As shown in~\cref{fig:teaser}, when fine-tuned on a 100K subset of LAION-COCO, our model achieves a \texttt{Comp} score of 53.5, significantly surpassing its pre-trained counterpart, and a \texttt{ZS} score of 55.9, nearly matching the pre-trained CLIP. Additionally, it attains an \texttt{I2T Ret} score of 58.2, the highest among models not fine-tuned on COCO.
Further improvements are observed with using LoRA~\cite{hu2022lora} for fine-tuning, which boosts the \texttt{Comp} score to 54.2 while maintaining the \texttt{ZS} score. Similar trends are evident when we fine-tune \texttt{FSC-CLIP} on a 100K subset of CC3M. Remarkably, these results are achieved by our innovative Local HN loss and Selective Calibrated Regularization (SCR) design. We further analyze these contributions in~\cref{sec:analysis}.

\begin{table*}[t]
\centering
\begin{subtable}[t]{0.32\linewidth}
    \centering
    \resizebox{.97\linewidth}{!}{%
    \begin{tabular}{cccccc}
    \toprule
    id & $\lambda_l$ &  \texttt{Comp}  &  \texttt{ZS}  &  \texttt{I2T Ret}  &  \texttt{T2I Ret}  \\
    \midrule
    1 &  - &    52.9    &      55.8      &   57.5    &   55.5    \\
    \cmidrule(lr){1-2} \cmidrule(lr){3-6}
    2 &  0.1 &    53.0    &      55.7      &   57.4    &   55.4    \\
    \rowcolor[rgb]{0.85,0.85,0.85}3 &  0.2 &    53.5    &      55.3      &   58.2    &   55.5    \\
    4 &  0.5 &    53.5    &      55.7      &   57.3    &   55.4    \\
    \bottomrule
    \end{tabular}%
    }
    \subcaption{Sensitivity to the weighting factor $\lambda_l$ of the local HN loss.}
    \label{tab:local_hn}
\end{subtable}
~
\begin{subtable}[t]{0.32\linewidth}
    \centering
    \resizebox{.97\linewidth}{!}{%
    \begin{tabular}{cccccc}
    \toprule
    id & $\gamma$ &  \texttt{Comp}  &  \texttt{ZS}  &  \texttt{I2T Ret}  &  \texttt{T2I Ret}  \\
    \midrule
    1 &  - &    53.9    &      53.8      &   51.7    &   54.9    \\
    \cmidrule(lr){1-2} \cmidrule(lr){3-6}
    2 &  1.0 &    53.4    &      54.9      &   54.7    &   55.1    \\
    \rowcolor[rgb]{0.85,0.85,0.85}3 &  2.0 &    53.5    &      55.3      &   58.2    &   55.5    \\
    4 &  5.0 &    52.3    &      55.6      &   60.2    &   55.5    \\
    \bottomrule
    \end{tabular}%
    }
    \subcaption{Sensitivity to the modulation factor $\gamma$ of focal loss.}
    \label{tab:focal_loss}
\end{subtable}
~
\begin{subtable}[t]{0.32\linewidth}
    \centering
    \resizebox{.99\linewidth}{!}{%
    \begin{tabular}{cccccc}
    \toprule
    id & $\beta$ &  \texttt{Comp}  &  \texttt{ZS}  &  \texttt{I2T Ret}  &  \texttt{T2I Ret}  \\
    \midrule
    1 &  - &    54.2    &      54.2      &   53.1    &   54.8    \\
    \cmidrule(lr){1-2} \cmidrule(lr){3-6}
    \rowcolor[rgb]{0.85,0.85,0.85}2 &  0.02 &    53.5    &      55.3      &   58.2    &   55.5    \\
    3 &  0.05 &    53.1    &      55.2      &   59.0    &   55.1    \\
    4 &  0.10 &    52.3    &      55.2      &   58.7    &   55.3    \\
    \bottomrule
    \end{tabular}%
    }
    \subcaption{Sensitivity to the label smoothing factor $\beta$.}
    \label{tab:label_smoothing}
\end{subtable}
\vspace{-1mm}
\caption{Sensitivity analysis of each component in our \texttt{FSC-CLIP} framework. \textbf{(a):} With the global HN loss applied, applying the local HN loss benefits the compositionality while preserving the multi-modal task scores. \textbf{(b)} and \textbf{(c)}: Both focal loss and label smoothing, the two components of our Selective Calibrated Regularization (SCR), mutually enhance multi-modal task performance but may compromise compositionality when applied too strongly. We highlight the cells corresponding to our design choices in the final \texttt{FSC-CLIP} model.}
\label{tab:sensitivity}
\end{table*}
\begin{table*}[t]
\centering
\begin{minipage}[t]{0.48\linewidth}
    \centering
    \resizebox{0.987\linewidth}{!}{%
    \begin{tabular}{lccccc}
    \toprule
    CLIP$^1$      & LoRA           &  \texttt{Comp}  &  \texttt{ZS}  &  \texttt{I2T Ret}  &  \texttt{T2I Ret}  \\
    \midrule
     ViT-B/16     &                &    46.2    &      60.3      &   62.9    &   49.0    \\
     \cmidrule(lr){1-2}\cmidrule(lr){3-6}
     ~+ NegCLIP   &                &    54.1    &      55.9      &   53.8    &   58.1    \\
     ~+ \textbf{\texttt{FSC-CLIP}}       &                &    54.1    &      57.0      &   59.7    &   59.3    \\
     ~+ \textbf{\texttt{FSC-CLIP}}       & \checkmark     &    54.6    &      57.4      &   59.9    &   58.8    \\
    \bottomrule
    \multicolumn{6}{l}{\small $^1$Pre-trained: 400M OpenAI, Fine-tuned: LAION-COCO 100K subset.} \\
    \end{tabular}%
    }
    \vspace{-1mm}
    \caption{Fine-tuning results of CLIP with a ViT-B/16 encoder, pre-trained on 400M samples of OpenAI data. 
    }
    \label{tab:vitb16}
\end{minipage}
~~~
\begin{minipage}[t]{0.48\linewidth}
    \centering
    \resizebox{0.99\linewidth}{!}{%
    \begin{tabular}{lccccc}
    \toprule
    CLIP$^2$      & LoRA           &  \texttt{Comp}  &  \texttt{ZS}  &  \texttt{I2T Ret}  &  \texttt{T2I Ret}  \\
    \midrule
     ViT-B/32        &                &    44.3    &      63.0      &   63.8    &   51.2    \\
     \cmidrule(lr){1-2}\cmidrule(lr){3-6}
     ~+ NegCLIP   &                &    53.5    &      59.2      &   52.1    &   52.3    \\
     ~+ \textbf{\texttt{FSC-CLIP}}       &                &    52.9    &      61.1      &   56.8    &   53.8    \\
     ~+ \textbf{\texttt{FSC-CLIP}}       & \checkmark     &    54.0    &      60.7      &   56.8    &   53.1    \\
    \bottomrule
    \multicolumn{6}{l}{\small $^2$Pre-trained: DataComp-XL, Fine-tuned: LAION-COCO 100K subset.} \\
    \end{tabular}%
}
    \vspace{-1mm}
    \caption{Fine-tuning results of CLIP with a ViT-B/32 encoder, pre-trained on 12.8B DataComp-XL.}
    \label{tab:datacomp}
\end{minipage}
\vspace{-3mm}
\end{table*}
\noindent \textbf{Robust Fine-tuning on Compositionality and Zero-shot Tasks.}
As depicted in~\cref{fig:wiseft}, we utilize the weight-space ensembling technique, WiSE-FT~\cite{wortsman2022robust}, to compare different fine-tuning methods and their trajectories, specifically in terms of \texttt{Comp} and \texttt{ZS} scores using LAION-COCO for fine-tuning our model. We create intermediate models by interpolating between each fine-tuned model and the pre-trained one. The blending ratio increases from 0.0 (\eg, pre-trained) to 1.0 (\eg, fully fine-tuned), in increments of 0.1.

\texttt{FSC-CLIP} with LoRA attains a \texttt{ZS} score of 58 at the intermediate, surpassing the scores of other models, while improving \texttt{Comp} to 50. When fully fine-tuned, it attains superior \texttt{Comp} score and offers better trade-offs than CLoVe and CE-CLIP, without significant loss in \texttt{ZS}.
In contrast, DAC-LLM sees a significant drop in \texttt{ZS}, gaining only 0.5 point in \texttt{Comp}, as highlighted by the red marker. Meanwhile, \texttt{FSC-CLIP} not only matches but exceeds the \texttt{ZS} score by 4.9 in the fully fine-tuned model.

\subsection{Analysis}
\label{sec:analysis}
We further present an in-depth analysis on our \texttt{FSC-CLIP}, fine-tuned on LAION-COCO: 

\noindent \textbf{Impact of Individual Components.}
From~\cref{tab:compo_analysis}, we observe that applying the local HN loss alone (row 2) surprisingly preserves the multi-modal scores. However, when both global and local HN losses are activated (row 3), \texttt{Comp} is further boosted but at the cost of \texttt{ZS} and \texttt{I2T Ret} scores, likely due to the complicated adverse effects of the losses. The proposed SCR effectively addresses this degradation. Both focal loss (row 4) and label smoothing (row 5) are effective and, when combined, complementarily boost all the \texttt{ZS}, \texttt{I2T Ret}, and \texttt{T2I Ret} scores. Notably, \texttt{I2T Ret} increases by 11.3 (rows 3 to 6) with only a relatively mild drop in \texttt{Comp}.
We also note that comparing rows 7 and 8 with rows 1 and 2, SCR significantly boosts multi-modal task scores. Furthermore, as shown in row 6, applying both global and local HN losses is essential for achieving better \texttt{Comp} and \texttt{I2T Ret} scores.

\noindent \textbf{Sensitivity Analysis.} 
We explore the impact of individually varying each component's parameters in the final model, as detailed in~\cref{tab:sensitivity}. From~\cref{tab:local_hn}, we find that increasing the local HN loss parameter $\lambda_l$ improves \texttt{Comp} score while preserving multi-modal task scores.
\cref{tab:focal_loss} shows that increasing the modulation parameter $\gamma$ boosts multi-modal tasks; however, beyond a certain point, we find that compositionality declines, as weakening the learning signal from HN texts. 
Similarly, \cref{tab:label_smoothing} indicates that label smoothing benefits multi-modal tasks, particularly \texttt{I2T Ret}. Yet, assigning too much positive margin with $\beta$ to negative samples can impede the learning of compositionality.

\noindent \textbf{Fine-tuning CLIP with ViT-B/16.}
We also fine-tuned CLIP with a ViT-B/16 encoder from OpenAI for comparison, as detailed in~\cref{tab:vitb16}. This model uses more image patches in training, showing better multi-modal capabilities. However, no gains are observed in \texttt{Comp} compared to the ViT-B/32 model from~\cref{tab:method_comparison}. After fine-tuning, NegCLIP decreases \texttt{ZS} and \texttt{I2T Ret} scores. In contrast, \texttt{FSC-CLIP} maintains its \texttt{Comp} score and significantly enhances multi-modal task performances. 
We also find that fine-tuning with LoRA yields improved \texttt{ZS} and \texttt{I2T Ret} scores, along with a higher \texttt{Comp} score.

\begin{figure*}[t]
  \centering
   \includegraphics[width=0.95\linewidth]{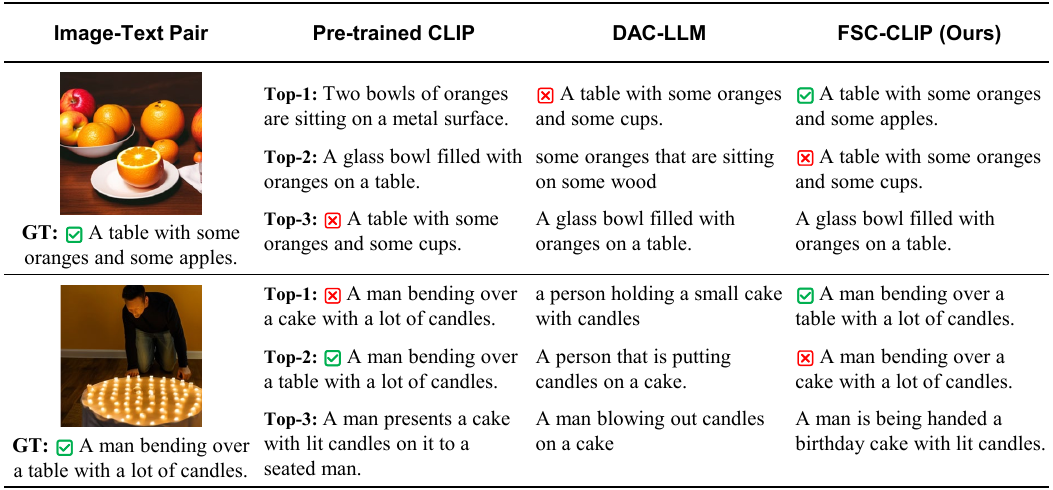}
   \vspace{-1mm}
   \caption{
Image to text retrieval examples on COCO-CF dataset. CLIP and DAC-LLM often rank negative captions (marked with red crossmarks) as top-1, while \texttt{FSC-CLIP} consistently retrieves the correct caption (marked with green checkmarks), demonstrating superior fine-grained understanding and retrieval accuracy in challenging conditions.
   }
   \label{fig:retrieval}
   \vspace{-2.5mm}
\end{figure*}
\noindent \textbf{Scaling Pre-training Data for Fine-tuning.}
We explore the effect of large-scale pre-training data when fine-tuned. From~\cref{tab:datacomp}, we fine-tuned a CLIP model with a ViT-B/32 encoder, pre-trained on 12.8B DataComp-XL dataset~\cite{gadre2024datacomp}, far exceeding the 400M samples from OpenAI~\cite{radford2021learning}.
Despite the larger scale pre-training yielding a promising \texttt{ZS} score of 63.0, we find no improvement in compositionality compared to OpenAI's CLIP.
For fine-tuning, NegCLIP results in a notable drop in multi-modal task performance. In contrast, \texttt{FSC-CLIP} with LoRA not only counters this degradation but also achieves a higher \texttt{Comp} score than NegCLIP. 

\noindent \textbf{Qualitative Counterfactual Image to Text Retrieval Results.}
In~\cref{fig:retrieval}, we compare image to text retrieval results on the COCO-Counterfactuals (COCO-CF)~\cite{le2024coco} dataset for three models: pre-trained CLIP~\cite{radford2021learning}, DAC-LLM~\cite{doveh2024dense}, and our proposed \texttt{FSC-CLIP}. The figure displays the top-3 retrieved captions for each image, with correct captions indicated by green checkmarks and incorrect ones by red crossmarks. We observe that CLIP and DAC-LLM often fail to retrieve the correct caption associated with the image, ranking a negative caption as top-1. In contrast, our \texttt{FSC-CLIP} consistently retrieves the correct caption as top-1, demonstrating superior retrieval capabilities along with a stronger fine-grained compositional understanding, even in the presence of hard negative captions.

\section{Conclusion}
In this paper, we introduce Fine-grained Selective Calibrated CLIP (\texttt{FSC-CLIP}), a new fine-tuning framework for vision-language compositionality. 
It aims to preserve multi-modal capabilities and address the limitations of existing methods relying on global representations. 
We achieve this by employing dense representations between images and texts and regularizing the hard negative losses to prevent degradation, thereby facilitating the introduction of Local Hard Negative Loss and Selective Calibrated Regularization. 
Our extensive validation shows improved compositional reasoning and promising performance in standard multi-modal tasks.

\noindent \textbf{Limitations.}
Our methodology, including all the prior approaches, relies on short captions for both training and evaluation benchmarks. This practice constrains the models' exposure to and understanding of longer contexts, which are essential for achieving a genuine vision-language compositional understanding. 
Longer and detailed captions involve more complex associations and contextual nuances~\cite{onoe2024docci,garg2024imageinwords} that are essential for advanced compositionality in vision and language models. Moving forward, there is a compelling need within the community to develop training and evaluation protocols that incorporate longer captions, better addressing the challenges of compositionality.

\noindent \textbf{Acknowledgements.}
This research was partially supported by Samsung Electronics Co., Ltd (G01200447), by the KAIST Cross-Generation Collaborative Lab Project, by the MSIT(Ministry of Science, ICT), Korea, under the Global Research Support Program in the Digital Field program(RS-2024-00436680) supervised by the IITP(Institute for Information \& Communications Technology Planning \& Evaluation), and by the Institute of Information and Communications Technology Planning and Evaluation (IITP) grant funded by the Korea Government (MSIT) (Artificial Intelligence Innovation Hub) under Grant 2021-0-02068. Additionally, this project was supported in part by Microsoft Research Asia. Dong-Jin Kim was supported by the National Research Foundation of Korea(NRF) grant funded by the Korea government(MSIT) (No. RS-2023-00245661).

\bibliography{custom}

\appendix
\begin{figure*}[t]
  \centering
   \includegraphics[width=0.94\linewidth]{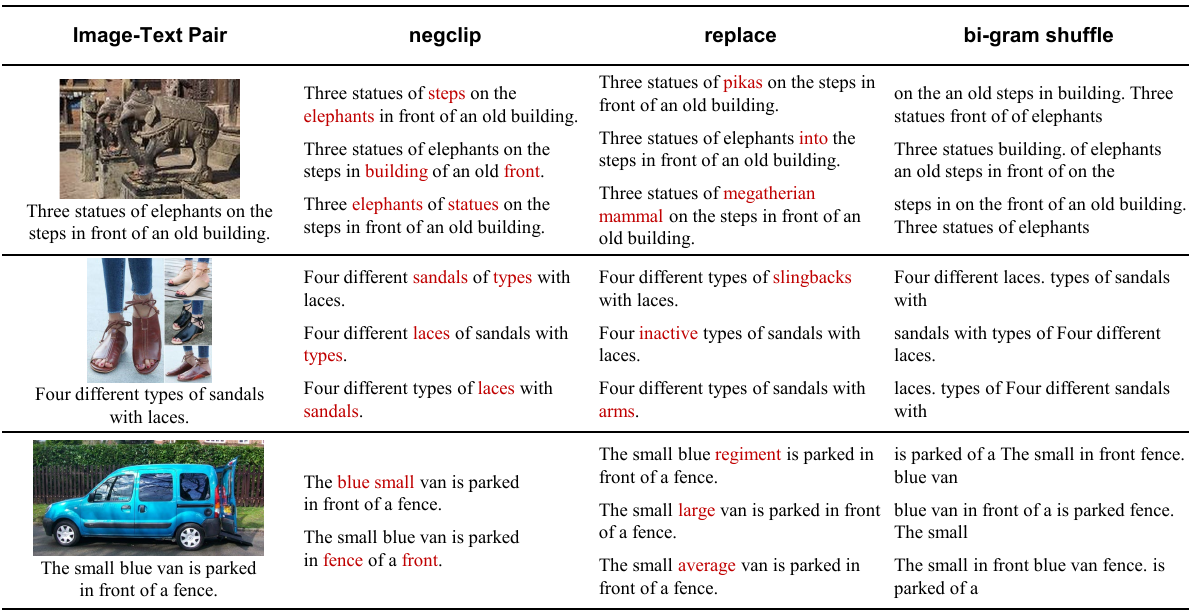}
   \caption{Example results of rule-based hard negative texts used for training our model. Image-text pairs were randomly sampled from LAION-COCO~\cite{Schuhmann2022laioncoco}. For \texttt{negclip}~\cite{yuksekgonul2023when} and \texttt{replace}~\cite{hsieh2024sugarcrepe}, differences from the original captions are highlighted in red.
   }
   \vspace{-2mm}
   \label{fig:neg_text_examples}
\end{figure*}

\section{Additional Details}
\subsection{Rule-based Hard Negative Texts}
\label{sec:sup_neg_texts}
We provide details in generating hard negative texts in our model. We employ three types of rule-based methods: \texttt{negclip}~\cite{yuksekgonul2023when}, \texttt{replace}~\cite{hsieh2024sugarcrepe}, and \texttt{bi-gram shuffle}. Each method is implemented in an online version and applied to the original text at every training step, resulting in total of four texts including the original caption for every batch as illustrated in~\cref{fig:overview}. In the online augmentation process, some captions do not yield a hard negative counterpart; these are masked out and excluded from the hard negative loss calculation.

The \texttt{negclip} method rearranges words within captions by swapping similar phrase types such as nouns, verbs, or adjectives within the text.

The \texttt{replace} method generates hard negative texts by replacing specific elements in the caption: entities, relations, or attributes, using antonyms or co-hyponyms from WordNet~\cite{fellbaum2010wordnet}.

The \texttt{bi-gram shuffle} rearranges text by shuffling bi-grams (\eg, pairs of adjacent words), within a sentence. It varies the sentence structure, ensuring the generated texts serve as challenging negatives to the original.

All the augmentation methods above utilize the SpaCy~\cite{spacy2} package. We implemented \texttt{bi-gram shuffle}, while for \texttt{negclip} and \texttt{replace}, we adopted the implementations from CLoVe~\cite{castro2024clove}. For illustrative purposes, we provide examples of each method applied to image-caption pairs, in~\cref{fig:neg_text_examples}.

\begin{table*}[ht]
\centering
\resizebox{.99\linewidth}{!}{%
    \begin{tabular}{p{3.5cm} p{1.8cm} p{5.5cm} p{8cm}}
    \toprule
        Benchmark  & License &  Image source & Tasks and Subtasks \\
    \midrule
    ARO & MIT & COCO, Visual Genome, Flickr30k & VG\_Relation, VG\_Attribution, Flickr30k\_Order, COCO\_Order \\
    CREPE-Productivity & \textit{unspecified} & Visual Genome & Atomic Foils, Negate, Swap \\
    SugarCrepe & MIT &  COCO & Add\_\{object, attribute\}, Replace\_\{object, attribute, relation\}, Swap\_\{object, attribute\}\\
    VALSE & MIT &  Visual7w, COCO, SWiG, VisDial\_v1.0, FOIL-it & Actions\_\{swap, replacement\}, Coreference\_\{hard, standard\}, Counting\_\{adversarial, hard, small\}, Existence, Foil-it, Plurals, Relations \\
    VL-Checklist & \textit{unspecified} &  Visual Genome, SWiG, COCO, HAKE, HICO\_Det, Pic, HCVRD, OpenImages & Object\_Location\_\{center, margin, mid\}, Object\_Size\_\{large, medium, small\}, Attribute\_\{action, color, material, size, state\}, Relation\_\{action, spatial\} \\
    WhatsUp & MIT & Controlled\_Images (\textit{self-captured}), COCO, GQA & Controlled\_Images\_\{A, B\}, COCO\_QA\_\{One, Two\}, VG\_QA\_\{One, Two\} \\
    \cmidrule(lr){1-4}
    ImageCoDe & MIT & OpenImages, MSRVTT, Video-Storytelling, YouCook & Static (\eg, images), Video (\eg, videos) \\
    SVO Probes & Apache-2.0 & Google Image Search API & Subject, Verb, Object \\
    \cmidrule(lr){1-4}
    Winoground & META IMAGES RESEARCH LICENSE & Getty Images & - \\ 
    EqBen & Apache-2.0 & Action Genome (AG), GEBC, YouCook2, Kubric, StableDiffusion (SD)  & EQ-AG, EQ-GEBC, EQ-YouCook2, EQ-Kubric\_\{location, counting, attribute\}, EQ-SD \\
    SPEC & \textit{unspecified} & Stable-Diffusion-XL 1.0 & Absolute\_size, Absolute\_position, Count, Relative\_size, Relative\_position, Existence \\
    \bottomrule
\end{tabular}
}
\vspace{-1mm}
\caption{A comprehensive list of compositionality benchmarks used in our work, further subdivided based on the query types for each individual test: \texttt{Image-to-Text}, \texttt{Text-to-Image}, and {Group}, respectively.
}
\label{tab:sup_benchmarks}
\vspace{-1mm}
\end{table*}

\subsection{Details on Evaluation Benchmark}
\label{sec:sup_details_eval}
\noindent \textbf{Compositionality.}
VLMs are presented with either an image or text query and must identify the correct match from a set of candidates, which includes subtly altered incorrect options of texts and images. If there are two candidates, including the original, the random chance accuracy becomes 0.5.

Benchmarks are grouped into three categories based on the query modality. \cref{tab:sup_benchmarks} provides a list of benchmarks for each category, along with the corresponding dataset licenses. 

\noindent (1) \texttt{Image-to-Text}, where the objective is to choose the correct textual description for a presented image: ARO~\cite{yuksekgonul2023when}, CREPE-Productivity~\cite{ma2023crepe}, SugarCrepe~\cite{hsieh2024sugarcrepe}, VALSE~\cite{parcalabescu-etal-2022-valse}, VL-Checklist~\cite{zhao-etal-2022-explainable}, and WhatsUp~\cite{kamath-etal-2023-whats}.

\noindent (2) \texttt{Text-to-Image}, which requires the selection of the correct image that matches a given text query: ImageCoDE~\cite{krojer-etal-2022-image} and SVO Probes~\cite{hendricks-nematzadeh-2021-probing}.

\noindent (3) \texttt{Group}, which involves two counterfactual image-text pairs, the challenge is to match each image with its corresponding text and the vice versa: Winoground~\cite{thrush2022winoground}, EqBen~\cite{wang2023equivariant}, and SPEC~\cite{peng2023synthesize}. 

For the \texttt{Image-to-Text} and \texttt{Text-to-Image} tasks, top-1 accuracy is used. For the \texttt{Group} tasks, group accuracy measures whether VLMs correctly match all the associated image-text pairs. 

To elaborate on details in specific benchmarks, for EqBen, we cap the evaluation sample size at 20,000. This is because the sub-tasks \texttt{eqbenag} and \texttt{eqbenyoucook2} contain 195,872 and 45,849 samples respectively, and evaluating all samples would be excessively time-consuming. Limiting the number of samples does not significantly alter the evaluation results. 
We do not use the official repository's 10\% evaluation split because it does not support sub-task-specific evaluations. 

For SVO-Probes, we have downloaded image-text pairs using the img2dataset~\cite{beaumont-2021-img2dataset} package from the URL list\footnote{\url{https://huggingface.co/datasets/MichiganNLP/svo_probes}}, as they are not available as physical files. 
Out of the original 36.8k samples, 22,162 were successfully downloaded, with 3,728 for the \texttt{subj\_neg}, 13,523 for the \texttt{verb\_neg}, and 4,911 for the \texttt{obj\_neg} sub-tasks, respectively.

For SPEC, unlike the other datasets in the \texttt{Group} category, we use the average of image to text and text to image accuracy, rather than group accuracy.

\noindent \textbf{Zero-shot Classification.} 
We leverage ELEVATER toolkit~\cite{li2022elevater} for 21 zero-shot classification tasks, including ImageNet~\cite{deng2009imagenet}, licensed under MIT License. 

\noindent \textbf{Image-Text Retrieval.} 
We utilize COCO captions~\cite{chen2015microsoft}, Flickr30k~\cite{young2014image}, and COCO-Counterfactuals~\cite{le2024coco} to evaluate the retrieval task. These datasets are licensed under BSD-3-Clause, CC0: Public Domain, and CC-BY-4.0, respectively.
For COCO-Counterfactuals, we randomly selected 30\% of the total 17,410 samples for evaluation, resulting in 5,223 samples. Each example includes two counterfactual image-text pairs, so the total number of images and texts used for retrieval is 10,446; one for the original and one for the hard negatives.

\subsection{Train Dataset}
We used the pre-processed version of COCO captions~\cite{chen2015microsoft} by~\citet{yuksekgonul2023when}, licensed under BSD 2-Clause. In addition, we utilized LAION-COCO~\cite{Schuhmann2022laioncoco}, licensed under CC-BY-4.0, and CC-3M~\cite{sharma-etal-2018-conceptual}\footnote{\url{https://github.com/google-research-datasets/conceptual-captions/blob/master/LICENSE}}, with 100K randomly sampled examples from each dataset to match the size of COCO for fine-tuning. We downloaded both datasets using the img2dataset package.  

\subsection{Baseline Methods}
In the comparisons with previous methods in~\cref{tab:method_comparison}, we evaluated prior approaches using the same protocol as ours to ensure fair and consistent evaluation. We obtained the corresponding checkpoints from each official repository and loaded them using the open\_clip package~\cite{ilharco_gabriel_2021_5143773}. 

When loading the checkpoints of previous models, we explicitly set \texttt{quick\_gelu} to True in the open\_clip implementation. 
While this setting was omitted in the implementations of NegCLIP~\cite{yuksekgonul2023when}, CE-CLIP~\cite{zhang2023contrasting}, and GNM-CLIP~\cite{sahin2024enhancing}, the adjustment aligns with the original CLIP models from~\cite{radford2021learning}, which were pre-trained and also fine-tuned with this option activated.

We list the previous methods with corresponding licenses. NegCLIP~\cite{yuksekgonul2023when}: MIT License, CE-CLIP~\cite{zhang2023contrasting}: MIT License, GNM-CLIP~\cite{sahin2024enhancing}: Apache-2.0 License, TSVLC\footnote{\url{https://github.com/SivanDoveh/TSVLC}} and DAC\footnote{\url{https://github.com/SivanDoveh/DAC}}~\cite{doveh2022teaching,doveh2024dense}: \textit{unspecified}, CLoVe~\cite{castro2024clove}: MIT License. 

\section{Additional Results}
For thoroughness, we include additional results not featured in the main paper. Note that all models were fine-tuned using the CLIP ViT-B/32 encoder from OpenAI~\cite{radford2021learning}.

\subsection{Additional Analysis}
\label{sec:sup_additional_analysis}
\begin{table}[t]
\centering
\resizebox{.95\linewidth}{!}{%
\begin{tabular}{crcccc}
\toprule
id & Attn. Norm. &  \texttt{Comp}  &  \texttt{ZS}  &  \texttt{I2T Ret}  &  \texttt{T2I Ret}  \\
\midrule
2 & minmax           &    51.7    &      \textbf{55.7}      &   \textbf{61.6}    &   54.5    \\
2 & minmax-sparse    &    51.6    &      55.5      &   61.1    &   \textbf{54.8}    \\
2 & softmax          &    \textbf{52.0}    &      55.4      &   60.9    &   54.6    \\
\cmidrule(lr){1-2} \cmidrule(lr){3-6}  
6 & minmax          &    \textbf{53.5}    &      55.3      &   \textbf{58.2}    &   55.5    \\
6 & minmax-sparse   &    53.4    &      55.1      &   57.8    &   55.4    \\
6 & softmax         &    53.3    &      \textbf{55.5}      &   57.1    &   \textbf{55.7}    \\
\bottomrule
\end{tabular}%
}
\caption{Ablation study on the normalization of attention weights in~\cref{eqn:attn_norm} for the LHN Loss. 
We found that no specific normalization method significantly impacted the results, highlighting the importance of the unique LHN loss design.}
\label{tab:sup_attn_norm}
\end{table}

\noindent \textbf{Normalization of attention weights.}
We present an ablation experiment on the normalization of attention weights in~\cref{eqn:attn_norm}, in alignment with the ablation study in~\cref{tab:compo_analysis}. We replace the current \texttt{minmax} normalization with \texttt{minmax-sparse}~\cite{bica2024improving} and \texttt{softmax}, respectively. As in \cref{tab:compo_analysis}, `id 2' only applies the LHN Loss without global HN loss and SCR, while `id 6' represents the full objective. Our findings show that the effectiveness of LHN Loss is not significantly impacted by any particular normalization technique. In other words, general normalization of attention weights can be applied to LHN Loss, reducing reliance on techniques like those from~\citet{bica2024improving}. This suggests that the unique design of LHN Loss is key to the improved performance.

\subsection{Multiple Runs}
In~\cref{tab:sup_multiple}, we report the mean and standard deviation for our models across all tasks listed in~\cref{tab:method_comparison}, using three distinct seeds: 0, 1, and 2 for training each model.

\subsection{Zero-shot Classification}
We report the results for each benchmark within the 21 zero-shot classification tasks in~\cref{tab:sup_eleveter}.
\begin{table*}[t]
\centering
\resizebox{.99\linewidth}{!}{%
\begin{tabular}{lcccccccccccccccc}
\toprule
Method & LoRA &  \rot{ARO}  &  \rot{CREPE}  &  \rot{EqBen}  &  \rot{ImageCoDe}  &  \rot{SugarCrepe}  &  \rot{SVO Probes}  &  \rot{VALSE}  &  \rot{VL-Checklist~~}  &  \rot{WhatsUp}  &  \rot{Winoground}  &  \rot{SPEC}  &  \rot{\texttt{Comp}}  &  \rot{\texttt{ZS}}  &  \rot{\texttt{I2T Ret}}  &  \rot{\texttt{T2I Ret}}  \\
\midrule

\multicolumn{17}{c}{\textit{Fine-tuned: LAION-COCO, 100K Samples}} \\
\textbf{\texttt{FSC-CLIP}} &      & 82.7$_{0.10}$ & 46.6$_{0.35}$ & 29.3$_{0.17}$ & 24.6$_{0.94}$  & 82.6$_{0.14}$   & 90.1$_{0.03}$   & 73.5$_{0.15}$ & 75.7$_{0.33}$     & 42.1$_{0.25}$ & 6.2$_{0.63}$    & 33.5$_{0.17}$ & 53.4$_{0.09}$ & 55.6$_{0.32}$     & 57.8$_{0.52}$ & 55.3$_{0.20}$ \\
\textbf{\texttt{FSC-CLIP}} & \checkmark & 
 85.3$_{0.14}$ & 52.9$_{1.28}$ & 28.9$_{0.17}$ & 24.9$_{0.11}$  & 80.5$_{0.11}$   & 89.7$_{0.05}$   & 72.4$_{0.17}$ & 78.7$_{0.20}$     & 42.9$_{0.05}$ & 5.4$_{0.38}$    & 32.4$_{0.11}$ & 54.0$_{0.17}$ & 56.1$_{0.18}$     & 57.3$_{0.13}$ & 54.4$_{0.08}$ \\
\bottomrule
\end{tabular}%
}
\caption{Evaluation across three training runs of our model using different seeds. 
We report the mean and standard deviation obtained from the evaluation results of the models across three trials.}
\label{tab:sup_multiple}
\end{table*}

\definecolor{mygray}{gray}{0.6}
\begin{table*}[t]
\centering
\resizebox{.99\linewidth}{!}{%
    \begin{tabular}{lcccccccccccccccccccccc}
\toprule
Method   &  \rot{caltech101}  &  \rot{cifar10}  &  \rot{cifar100}  &  \rot{country211}  &  \rot{dtd}  &  \rot{eurosat-clip}  &  \rot{fer2013}  &  \rot{fgvc-aircraft-2013b}  &  \rot{flower102}  &  \rot{food101}  &  \rot{gtsrb}  &  \rot{hateful-memes}  &  \rot{imagenet-1k}  &  \rot{kitti-distance}  &  \rot{mnist}  &  \rot{oxford-iiit-pets}  &  \rot{patchcamelyon}  &  \rot{rendered-sst2}  &  \rot{resisc45-clip}  &  \rot{stanfordcar}  &  \rot{voc2007classification~~}  &  \rot{Average}  \\
\midrule
CLIP-ViT-B/32 &     88.3     &   89.8    &    65.1    &     17.2     & 44.4  &      45.5      &   42.3    &         19.7          &    66.7     &   84.0    &  32.6   &      55.9       &     63.3      &       27.4       &  48.3   &        87.1        &      60.6       &      58.6       &      60.0       &     59.7      &          82.6           &   57.1    \\
\cmidrule(lr){1-1} \cmidrule(lr){2-22} \cmidrule(lr){14-17} \cmidrule(lr){23-23}
\multicolumn{23}{c}{\textit{Fine-tuned: MS-COCO, 100K Samples}} \\
NegCLIP  &     88.2     &   88.9    &    63.2    &     15.0     & 43.1  &      47.3      &   47.6    &         16.8          &    62.3     &   79.4    &  30.2   &      54.3       &     60.9      &       27.6       &  49.7   &        85.4        &      59.7       &      58.8       &      56.9       &     54.0      &          84.4           &   55.9    \\
CE-CLIP  &     82.2     &   85.9    &    60.2    &     9.6      & 35.2  &      44.9      &   39.7    &         10.0          &    47.2     &   70.1    &  28.0   &      53.5       &     49.9      &       34.6       &  40.6   &        66.0        &      58.8       &      61.1       &      51.5       &     35.3      &          83.1           &   49.9    \\
GNM-CLIP                                           &     86.8     &   88.4    &    65.7    &     15.2     & 42.0  &      50.1      &   46.6    &         17.3          &    62.4     &   81.8    &  30.2   &      54.9       &     61.4      &       25.2       &  54.4   &        86.3        &      59.0       &      58.5       &      58.7       &     53.1      &          84.0           &   56.3    \\
\cmidrule(lr){1-1} \cmidrule(lr){2-22} \cmidrule(lr){14-17} \cmidrule(lr){23-23}
\multicolumn{23}{c}{\textit{Fine-tuned: Conceptual Captions -- 3M (CC-3M), 100K Samples}} \\
TSVLC (RB) &     83.7     &   92.3    &    66.0    &     16.2     & 39.5  &      52.1      &   43.6    &         14.7          &    58.2     &   81.2    &  24.2   &      57.8       &     58.5      &       30.4       &  46.9   &        85.5        &      50.0       &      59.8       &      58.6       &     49.2      &          84.7           &   54.9    \\
TSVLC (RB+LLM)&     84.6     &   92.0    &    66.8    &     16.2     & 40.3  &      56.5      &   46.8    &         13.8          &    58.5     &   81.6    &  27.1   &      56.9       &     59.7      &       27.8       &  43.9   &        84.7        &      50.5       &      60.1       &      59.5       &     50.5      &          84.7           &   55.4    \\
DAC-LLM  &     82.6     &   90.4    &    64.1    &     14.3     & 38.4  &      52.5      &   50.7    &         10.5          &    49.7     &   74.1    &  24.2   &      56.3       &     51.0      &       16.3       &  42.1   &        74.4        &      50.0       &      54.5       &      52.2       &     39.4      &          85.1           &   51.1    \\
DAC-SAM &     81.3     &   89.9    &    64.1    &     14.8     & 40.4  &      49.8      &   48.0    &          8.9          &    48.9     &   72.3    &  24.9   &      55.7       &     52.3      &       18.7       &  45.2   &        76.7        &      58.9       &      60.0       &      54.7       &     39.8      &          84.1           &   51.9    \\
\cmidrule(lr){1-1} \cmidrule(lr){2-22} \cmidrule(lr){14-17} \cmidrule(lr){23-23}
\multicolumn{23}{c}{\textit{Fine-tuned: LAION-COCO, 600M Samples}} \\
CLoVe &     85.5     &   85.8    &    66.2    &     12.6     & 37.7  &      49.1      &   38.0    &          9.0          &    44.6     &   71.9    &  22.6   &      54.6       &     53.1      &       34.9       &  36.4   &        74.2        &      56.7       &      51.3       &      55.2       &     48.7      &          81.9           &   51.0    \\
\cmidrule(lr){1-1} \cmidrule(lr){2-22} \cmidrule(lr){14-17} \cmidrule(lr){23-23}
\multicolumn{23}{c}{\textit{Fine-tuned: LAION-COCO, 100K Samples}} \\
\textbf{\texttt{FSC-CLIP} (Ours)}      &     86.5     &   87.5    &    65.7    &     15.3     & 42.4  &      43.9      &   48.9    &         14.9          &    55.5     &   80.5    &  31.6   &      55.9       &     58.1      &       29.1       &  52.4   &        84.2        &      61.0       &      56.0       &      56.9       &     52.0      &          83.6           &   55.3    \\
\textbf{\texttt{FSC-CLIP} (Ours, LoRA)} &     85.9     &   88.5    &    66.3    &     15.8     & 39.8  &      52.8      &   48.2    &         14.2          &    57.0     &   81.0    &  27.9   &      56.3       &     57.4      &       33.9       &  54.3   &        82.7        &      59.8       &      57.2       &      58.7       &     52.6      &          83.7           &   55.9    \\
\bottomrule
\end{tabular}%
}
\caption{Expanded results for the 21 zero-shot classification tasks from ELEVATER~\cite{li2022elevater}.}
\label{tab:sup_eleveter}
\vspace{-1mm}
\end{table*}

\begin{table*}[t]
\centering
\resizebox{.99\linewidth}{!}{%
\begin{tabular}{lcccccccccccccccccccc}
\toprule
 & \multicolumn{6}{c}{COCO Retrieval} & \multicolumn{6}{c}{Flickr30k Retrieval} & \multicolumn{6}{c}{COCO-Counterfactuals Retrieval} & \multicolumn{2}{c}{Avg.} \\ 
 & \multicolumn{3}{c}{Image to text (I2T)} & \multicolumn{3}{c}{Text to image (T2I)} & \multicolumn{3}{c}{Image to text (I2T)} & \multicolumn{3}{c}{Text to image (T2I)} & \multicolumn{3}{c}{Image to text (I2T)} & \multicolumn{3}{c}{Text to image (T2I)} & I2T & T2I \\
 Method &  R@1  &  R@5  &  R@10  &  R@1  &  R@5  &  R@10  &  R@1  &  R@5  &  R@10  &  R@1  &  R@5  &  R@10  &  R@1  &  R@5  &  R@10  &  R@1  &  R@5  &  R@10  &  R@1  &  R@1  \\
\midrule
 CLIP-ViT-B-32                                      &   50.1    &   74.9    &    83.5    &   30.4    &   56.0    &    66.8    &   78.8    &   94.9    &    98.3    &   58.7    &   83.5    &    90.0    &   51.0    &   79.3    &    86.7    &   48.1    &   77.4    &    85.9    &   60.0    &   45.8    \\
\cmidrule(lr){1-1} \cmidrule(lr){2-7} \cmidrule(lr){8-13} \cmidrule(lr){14-19} \cmidrule(lr){20-21} 
\multicolumn{21}{c}{\textit{Fine-tuned: MS-COCO, 100K Samples}} \\
 NegCLIP                                            &   59.3    &   82.8    &    89.4    &   45.2    &   72.1    &    81.7    &   85.7    &   96.4    &    98.8    &   71.6    &   91.8    &    95.7    &   55.3    &   82.5    &    89.2    &   58.3    &   84.9    &    91.3    &   66.8    &   58.4    \\
 CE-CLIP                                            &   56.0    &   81.6    &    89.0    &   47.1    &   74.1    &    83.1    &   75.3    &   93.2    &    96.9    &   68.9    &   89.6    &    94.2    &   46.3    &   75.7    &    84.5    &   56.2    &   83.6    &    90.5    &   59.2    &   57.4    \\
 GNM-CLIP                                           &   58.1    &   81.4    &    88.8    &   41.1    &   67.5    &    77.8    &   82.9    &   96.2    &    98.6    &   68.8    &   89.9    &    94.1    &   57.2    &   84.5    &    90.5    &   56.7    &   84.5    &    91.1    &   66.1    &   55.5    \\
\cmidrule(lr){1-1} \cmidrule(lr){2-7} \cmidrule(lr){8-13} \cmidrule(lr){14-19} \cmidrule(lr){20-21} 
\multicolumn{21}{c}{\textit{Fine-tuned: Conceptual Captions -- 3M (CC-3M), 100K Samples}} \\
 TSVLC (RB)                                           &   46.1    &   71.7    &    80.4    &   36.3    &   62.0    &    72.4    &   74.0    &   93.2    &    96.4    &   64.9    &   87.2    &    92.7    &   44.6    &   72.0    &    80.2    &   55.0    &   83.3    &    90.0    &   54.9    &   52.1    \\
 TSVLC (RB+LLM)                                       &   46.4    &   71.8    &    80.8    &   36.6    &   62.2    &    72.7    &   74.8    &   92.6    &    96.8    &   65.1    &   87.6    &    92.7    &   44.1    &   71.5    &    80.1    &   55.1    &   83.3    &    90.4    &   55.1    &   52.3    \\
 DAC-LLM                                            &   29.9    &   54.5    &    65.6    &   37.3    &   63.5    &    73.8    &   52.9    &   79.8    &    87.9    &   64.6    &   88.0    &    93.0    &   28.1    &   53.6    &    64.4    &   55.2    &   83.0    &    90.0    &   36.9    &   52.4    \\
 DAC-SAM                                            &   33.1    &   57.9    &    68.8    &   34.0    &   59.7    &    70.0    &   59.8    &   82.7    &    89.0    &   61.7    &   85.7    &    91.2    &   30.4    &   55.2    &    64.8    &   51.5    &   79.9    &    87.3    &   41.1    &   49.0    \\
 \cmidrule(lr){1-1} \cmidrule(lr){2-7} \cmidrule(lr){8-13} \cmidrule(lr){14-19} \cmidrule(lr){20-21} 
\multicolumn{21}{c}{\textit{Fine-tuned: LAION-COCO, 600M Samples}} \\
 CLoVe                                              &   48.3    &   73.9    &    82.8    &   42.7    &   68.7    &    78.2    &   69.5    &   90.4    &    95.6    &   68.7    &   90.0    &    94.5    &   41.5    &   69.1    &    78.3    &   56.5    &   84.2    &    90.8    &   53.1    &   56.0    \\
\cmidrule(lr){1-1} \cmidrule(lr){2-7} \cmidrule(lr){8-13} \cmidrule(lr){14-19} \cmidrule(lr){20-21} 
\multicolumn{21}{c}{\textit{Fine-tuned: LAION-COCO, 100K Samples}} \\
 \textbf{\texttt{FSC-CLIP} (Ours)}      &   49.7    &   73.6    &    82.4    &   40.4    &   66.4    &    76.4    &   75.6    &   93.3    &    97.4    &   68.2    &   90.0    &    94.3    &   49.2    &   77.5    &    85.8    &   57.9    &   85.4    &    91.4    &   58.2    &   55.5    \\
 \textbf{\texttt{FSC-CLIP} (Ours, LoRA)} &   48.2    &   73.6    &    81.8    &   39.0    &   64.9    &    75.0    &   75.1    &   93.2    &    96.4    &   66.9    &   88.6    &    93.6    &   48.5    &   76.0    &    84.4    &   57.1    &   84.7    &    91.0    &   57.3    &   54.3    \\
\bottomrule
\end{tabular}%
}
\caption{Expanded results for the three zero-shot image-text retrieval tasks, including COCO~\cite{chen2015microsoft}, Flickr30k~\cite{young2014image}, and COCO-Counterfactuals~\cite{le2024coco}.}
\label{tab:sup_retrieval}
\vspace{-1mm}
\end{table*}

\subsection{Image-Text Retrieval}
We present the results for each benchmark included in the three image-text retrieval tasks in~\cref{tab:sup_retrieval}.

\end{document}